\documentclass[11pt,a4paper]{article}
\usepackage{authblk}
\usepackage{stackengine}
\usepackage[hyperref]{styles}
\usepackage{times}
\usepackage{latexsym}

\usepackage{url}
\usepackage{xcolor}
\usepackage{upgreek}
\usepackage[shortlabels]{enumitem}
\usepackage{tabu}
\setlist[enumerate]{nosep}
\usepackage{floatrow}
\usepackage{multirow}
\usepackage{makecell}
\usepackage{amsmath}

\usepackage{booktabs}
\usepackage{graphicx}
\usepackage{amsfonts}
\usepackage{tabularx}
\usepackage{bm}
\aclfinalcopy

\usepackage{microtype}

\def\aclpaperid{***} 

\newcommand*{\MyIndent}{\hspace*{0.5cm}}

\title{Conversational Question Answering over Knowledge Graphs with Transformer and Graph Attention Networks}

\author[1]{Endri Kacupaj}
\author[2]{Joan Plepi}
\author[3]{Kuldeep Singh}
\author[4]{Harsh Thakkar}
\author[1,5]{\\Jens Lehmann}
\author[1]{Maria Maleshkova}

\affil[1]{Smart Data Analytics Group, University of Bonn, Bonn, Germany}

\affil[2]{Technische Universität Darmstadt, Darmstadt, Germany}

\affil[3]{Zerotha Research and Cerence GmbH, Germany}

\affil[4]{Zerotha Research and Osthus GmbH, Germany}

\affil[5]{Fraunhofer IAIS, Dresden, Germany\vspace{0.5em}}

\affil[ ]{
\ttfamily{\{kacupaj,jens.lehmann,maleshkova\}@cs.uni-bonn.de} \protect\\
\ttfamily{joan.plepi@tu-darmstadt.de} \protect\\
\ttfamily{kuldeep.singh1@cerence.com} \protect\\
\ttfamily{harsh.thakkar@osthus.com} \protect\\
\ttfamily{jens.lehmann@iais.fraunhofer.de}
}

\date{}

\begin{document}
\maketitle
\vspace*{6em}
\begin{abstract}
This paper addresses the task of (complex) conversational question answering over a knowledge graph. For this task, we propose LASAGNE (mu\textbf{L}ti-task sem\textbf{A}ntic par\textbf{S}ing with tr\textbf{A}nsformer and \textbf{G}raph atte\textbf{N}tion n\textbf{E}tworks). It is the first approach, which employs a transformer architecture extended with Graph Attention Networks for multi-task neural semantic parsing. LASAGNE uses a transformer model for generating the base logical forms, while the Graph Attention model is used to exploit correlations between (entity) types and predicates to produce node representations. LASAGNE also includes a novel entity recognition module which detects, links, and ranks all relevant entities in the question context. We evaluate LASAGNE on a standard dataset for complex sequential question answering, on which it outperforms existing baseline averages on all question types.  Specifically, we show that LASAGNE improves the F1-score on eight out of ten question types; in some cases, the increase in F1-score is more than 20\% compared to the state of the art. 
\end{abstract}

\section{Introduction} \label{sec:introduction}
\begin{figure}[!t]
\vspace*{6em}
\centering
\captionsetup{type=figure}
\includegraphics[width=0.85\textwidth]{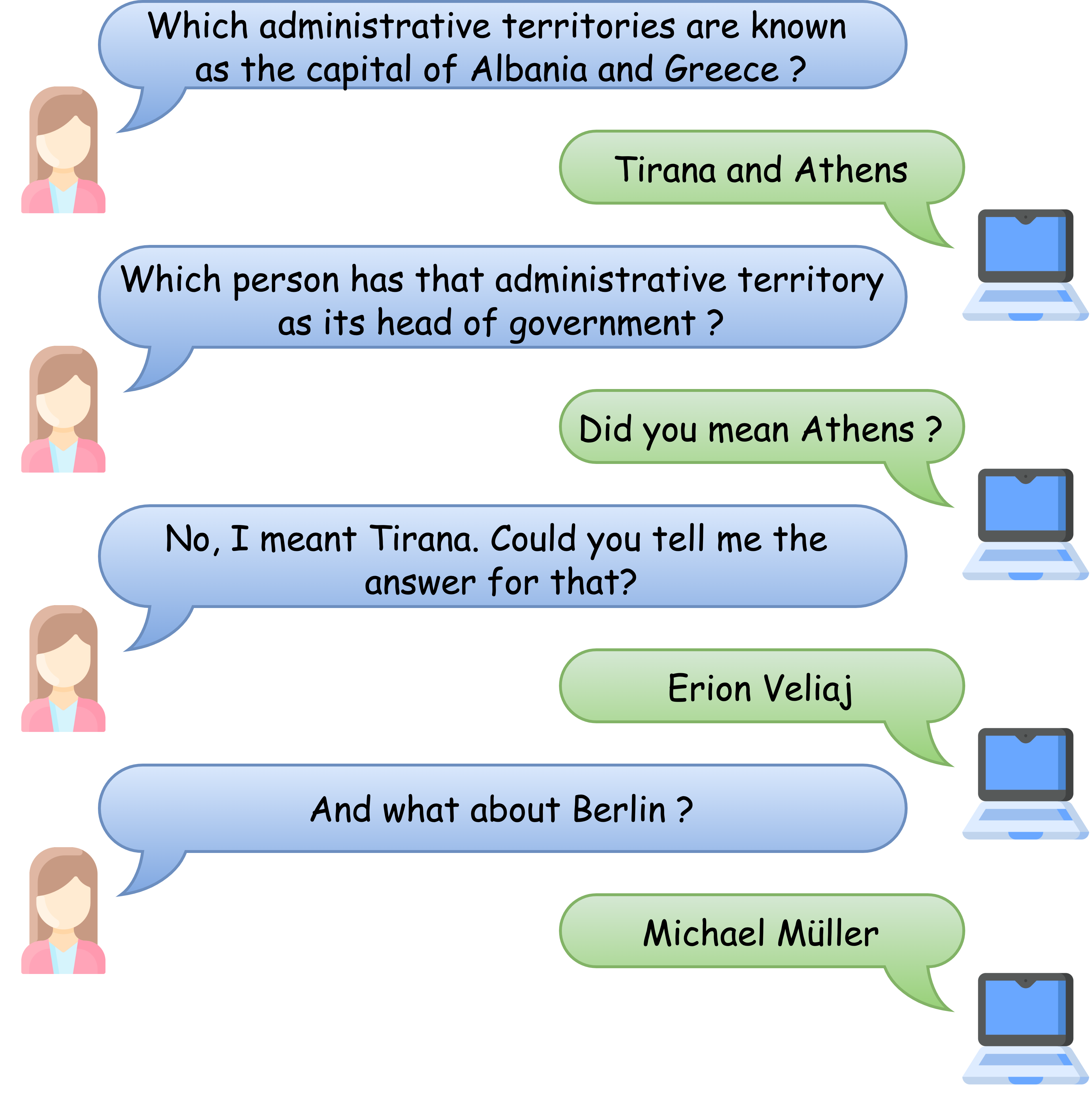}
\caption{Conversational Question Answering task with examples similar to CSQA dataset~\cite{saha2018complex}.}
\label{fig:chat_examples}
\end{figure}

Since their inception in the late 2000s, publicly available Knowledge Graphs (e.g., DBpedia~\cite{madoc37476} and Yago~\cite{suchanek2007yago}) have been widely used as a source of knowledge in several natural language processing (NLP) tasks such as entity linking, relation extraction, fact-checking, and question answering. Question answering (QA), in particular, is an essential task that maps a user natural language question to a query over a knowledge graph (KG) to retrieve the correct answer~\cite{singh2018reinvent}. With the increasing popularity of intelligent personal assistants (e.g., Alexa, Siri), the research focus has been shifted to conversational question answering that involves multi-turn dialogues, incorporating the phenomenon of anaphora and ellipses
~\cite{christmann2019look,shen-etal-2019-multi}(c.f. Figure~\ref{fig:chat_examples}). 

Conversational QA is often realised by using semantic parsing approaches, mapping an utterance to a logic form for extracting answers from a KG ~\cite{guo2018dialog,shen-etal-2019-multi}. The state of the art for semantic parsing approaches decomposes the semantic parsing process into two stages \cite{shen-etal-2019-multi}. First, a logical form is generated based on low-level features and then the missing details are filled by considering both the question and the template. Other approaches
~\cite{dong-lapata-2016-language, guo2018dialog, liang2017neural} first employ an entity linking model to identify entities in the question and subsequently use another model to map the question to a logical form.  \cite{zhang2018variational, shen-etal-2019-multi} point out that the modular approaches suffer from the common issue of error propagation along the QA pipeline, resulting in accumulated errors. To mitigate these errors, Shen et al.~\shortcite{shen-etal-2019-multi} proposed a multi-task framework, where a pointer-equipped semantic parsing model was designed to resolve coreference in conversations and empower joint learning with a type-aware entity detection model. Furthermore, the authors used simple classifiers to predict the required (entity) types and predicates for the generated logical forms. In this paper, we argue that Shen et al.~\shortcite{shen-etal-2019-multi} model (the current SotA) has the following shortcomings:  1) the (entity) type and predicate classifiers share no common information, except for the supervision signal propagated to them. 2) Hence, due to missing common information, the model can produce ambiguous results, since the classifiers can predict entities and predicates that do not correlate with each other.

\textbf{Approach and Contributions}: 
We tackle the problem of conversational (complex) question answering over a large-scale knowledge graph. We propose LASAGNE (mu\textbf{L}ti-task sem\textbf{A}ntic par\textbf{S}ing with tr\textbf{A}nsformer and \textbf{G}raph atte\textbf{N}tion n\textbf{E}tworks) - a multi-task learning framework consisting of a transformer model extended with Graph Attention Networks (GATs)~\cite{velickovic2018graph} for multi-task neural semantic parsing. Our framework handles semantic parsing using the transformer~\cite{vaswani2017attention} model similar to previous approaches. However, in LASAGNE we introduce the following two novel contributions: 1) the transformer model is supplemented with a Graph Attention Network to exploit the correlations between (entity) types and predicates due to its message-passing ability between the nodes. 2) We propose a novel entity recognition module that detects, links, filters, and permutes all relevant entities. \cite{shen-etal-2019-multi}
~uses a pointer equipped decoder that learns and identifies the relevant entities for the logical form using only the encoder's information. In contrast, we use both sources of information, i.e., the entity detection module and the encoder, to filter and permute the relevant entities for a logical form. This avoids re-learning entity information in the current question context and relies on the entity detection module's information. Our empirical results show that the proposed novel contributions lead to substantial performance improvements. 

LASAGNE achieves the state of the art results in 8 out of 10 question types on the Complex Sequential Question Answering (CSQA)~\cite{saha2018complex} dataset consisting of conversations over linked QA pairs. The dataset contains 200K dialogues with 1.6M turns, and over 12.8M entities from Wikidata\footnote{\url{https://www.wikidata.org/}}. Our implementation, the annotated dataset with the proposed grammar, and the results are publicly available to facilitate reproducibility and reuse\footnote{\url{https://github.com/endrikacupaj/LASAGNE}}.

The structure of the paper is as follows: Section~\ref{sec:related_work} summarises the related work.   Section~\ref{sec:model} presents the proposed LASANGE framework. Section~\ref{sec:experiment} describes the experiments, including the experimental setup, the results, the ablation study and error analysis. We conclude in Section~\ref{sec:conclusion}.

\begin{table*}[!ht]
\small%
\centering
\resizebox{\textwidth}{!}{%
\begin{tabular}{lp{11cm}}
\toprule
\textbf{Action} & \textbf{Description} \\
\hline
set $\rightarrow$ find(\textit{e}, \textit{p}) & set of objects part of the triples with subject \textit{e} and predicate \textit{p}   \\
set $\rightarrow$ find\_reverse(\textit{e}, \textit{p}) &  set of subjects part of the triples with object \textit{e} and predicate \textit{p}\\
set $\rightarrow$ filter\_type(set, tp) & filter the given set of entities based on the given type \\
set $\rightarrow$ filter\_multi\_types($set_1$, $set_2$) & filter the given set of entities based on the given set of types \\ 
dict $\rightarrow$ find\_tuple\_counts(p, $tp_1$, $tp_2$) & extracts a dictionary, where keys are entities of $type_1$ and values are the number of objects of $type_2$ related with \textit{p} \\
dict $\rightarrow$ find\_reverse\_tuple\_counts(p, $tp_1$, $tp_2$) & extracts a dictionary, where keys are entities of $type_1$ and values are the number of subjects of $type_2$ related with \textit{p} \\
set $\rightarrow$ greater(dict, num) & set of those entities that have greater count than \textit{num} \\
set $\rightarrow$ lesser(dict, num) & set of those entities that have lesser count than \textit{num} \\
set $\rightarrow$ equal(dict, num) & set of those entities that have equal count with \textit{num} \\
set $\rightarrow$ approx(dict, num) & set of those entities that have approximately same count with \textit{num} \\
set $\rightarrow$ atmost(dict, num) & set of those entities that have at most same count with \textit{num} \\
set $\rightarrow$ atleast(dict, num) & set of those entities that have at least same count with \textit{num} \\
set $\rightarrow$ argmin(dict) & set of those entities that have the most count \\
set $\rightarrow$ argmax(dict) & set of those entities that have the least count \\
boolean $\rightarrow$ is\_in(entity, set) & check if the entity is part of the set  \\
number $\rightarrow$ count(set) & count the number of elements in the set \\
set $\rightarrow$ union($set_1$, $set_2$) & union of $set_1$ and $set_2$\\
set $\rightarrow$ intersection($set_1$, $set_2$) & intersection of $set_1$ and $set_2$ \\
set $\rightarrow$ difference($set_1$, $set_2$) &  difference of $set_1$ and $set_2$ \\
\bottomrule
\end{tabular}
}
\caption{Predefined grammar with respective actions to generate logical forms.}
\label{tab:grammar}
\end{table*}

\section{Related Work} \label{sec:related_work}
We point to the survey by~\cite{gao2018neural} that provides a holistic overview of neural approaches in conversational AI. In this paper, we stick to our closely related work, i.e., semantic parsing-based approaches in conversations. \cite{liang2017neural} introduce a neural symbolic machine (NSM) extended with a key-value memory network, where keys and values are the output of the sequence model in different encoding or decoding steps. The NSM model is trained using the REINFORCE algorithm with weak supervision and evaluated on the WebQuestionsSP dataset \cite{yih2016value}.

\cite{saha2018complex} propose a hybrid model of the HRED model \cite{serban2016building} and the key-value memory network model \cite{miller2016key}. The model consists of three components. The first one is the Hierarchical Encoder, which computes a representation for each utterance. The next module is a higher-level encoder that computes a representation for the context. The second component is the Key-Value Memory Network. It stores each of the candidate tuples as a key-value pair where the key contains the concatenated embedding of the relation and the subject. In contrast, the value contains the embedding of the object. The last component is the decoder used to create an end-to-end solution and produce multiple types of answers.

\cite{guo2018dialog} present a model that converts an utterance in conversation to a logical form. The model follows a flexible grammar, in which the generation of a logical form is equivalent to predicting a sequence of actions. A dialogue memory management is proposed and integrated into the model, so that historical entities, predicates, and action sub-sequences can selectively be replicated. \cite{shen-etal-2019-multi} proposed the first multi-task learning framework that learns type-aware entity detection and pointer-equipped logical form generation simultaneously. The multi-task learning framework takes advantage of the supervision from the subtasks.

\begin{figure*}[!ht]
\includegraphics[width=0.8\textwidth]{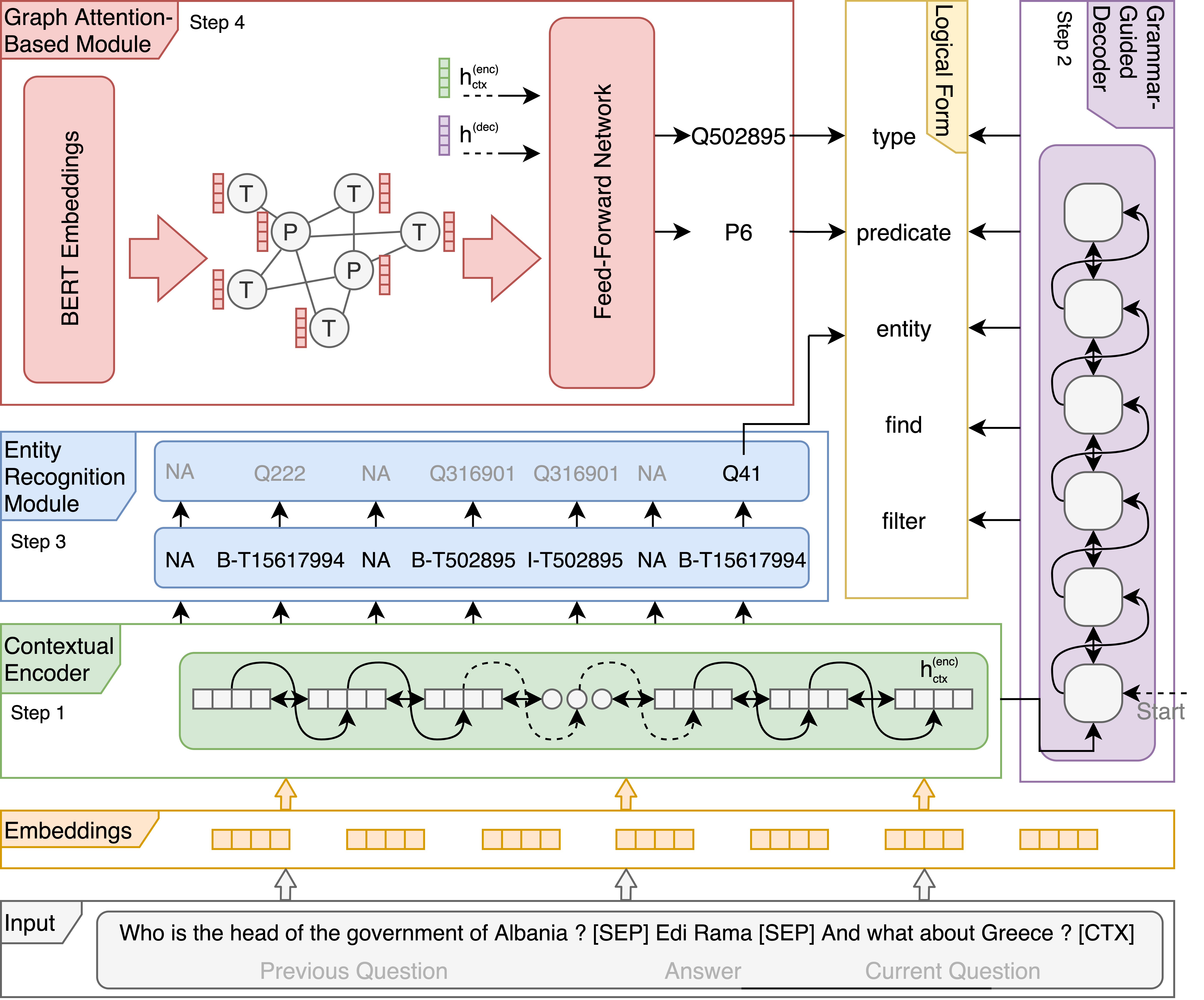}
\caption{LASAGNE (Multi-task Semantic Parsing with Transformer and Graph Attention Networks) architecture. It consists of three modules: 1) A semantic parsing-based transformer model, containing a contextual encoder and a grammar guided decoder using the grammar defined in Table~\ref{tab:grammar}. 2) An entity recognition module, which identifies all the entities in the context, together with their types, linking them to the knowledge graph. It filters them based on the context and permutes them, in case of more than one required entity. Finally, 3) a graph attention-based module that uses a GAT network initialised with BERT embeddings to incorporate and exploit correlations between (entity) types and predicates. The resulting node embeddings, together with the context hidden state ($h_{ctx}$) and decoder hidden state ($d_h$), are used to score the nodes and predict the corresponding type and predicate.}
\label{fig:model_architecture}
\end{figure*}

\section{LASAGNE}\label{sec:model}
In a conversation, the input data consists of utterances $u$ and their answers $a$, extracted from the knowledge graph. Our framework LASAGNE employs a multi-task semantic parsing approach. In particular, it maps the utterance $u$ to a logical form $z$, depending on the conversation context. 
Figure~\ref{fig:model_architecture} shows the architecture of LASAGNE.

\subsection{Grammar}

For the semantic parsing task, we propose a grammar that can be used to capture the entire context of the input utterance with the minimum number of actions. Table~\ref{tab:grammar} illustrates the complete grammar with all the defined actions. We considered the work by \cite{guo2018dialog} as a starting point for generating them, however, we have updated many of the semantic actions. For instance, for a couple of actions, we also define their reverse occurrence (e.g. \emph{find}, \emph{find\_reverse)}).

\subsection{Transformer}
To translate the input conversation into a sequence of actions (logical form), we utilise a transformer model \cite{vaswani2017attention}. 
Specifically, the transformer here aims to map a question $q$, that is a sequence $x = \{x_1,\dots, x_n\}$, to the answer label $l$, that can be also defined as a sequence $y = \{y_1,\dots, y_m\}$, by modelling the conditional probability $p(y | x)$.

\subsubsection{Input and Word Embedding}
We have to incorporate the dialog history from previous interactions as an additional input to our model for handling coreference and ellipsis. To do so, we consider the following utterances for each turn: 1) the previous question, 2) the previous answer, and 3) the current question. Utterances are separated from one another by using a $[SEP]$ token. At the end of the last utterance, we append a context token $[CTX]$, which is used as the semantic representation for the entire input question. 
In the next step, given an utterance $q$ containing $n$ words $\{w_1,\dots,w_n\}$ we first tokenise the conversation context using WordPiece tokenization \cite{2016arXiv160908144W}, and after that, we use the pre-trained model GloVe \cite{pennington-etal-2014-glove} to embed the words into a vector representation space of dimension $d$ \footnote{Across the model, we use the same dimension $d$ for all the representations, unless it is explicitly noted.}. Our word embedding model provides us with a sequence $x = \{x_1,\dots,x_n\}$ where $x_i$ is given by, $x_i = GloVe(w_i)$ and $x_i \in \mathbb{R}^{d}$.

\subsubsection{Contextual Encoder}
The word embeddings $x$, are forwarded as input to the contextual encoder, which uses the multi-head attention mechanism described by \cite{vaswani2017attention}. The encoder here outputs the contextual embeddings $h^{(enc)} =  \{h_{1}^{(enc)},\dots,h_{n}^{(enc)}\}$, where $h_{i}^{(enc)} \in \mathbb{R}^{d}$ and it can be defined as: 
\begin{equation}
\begin{split}
    &h^{(enc)} = encoder(x; \theta^{(enc)}),
\end{split}
\end{equation}
\noindent where $\theta^{(enc)}$ are the encoder's trainable parameters.
\subsubsection{Grammar-Guided Decoder}
We use a grammar guided decoder for generating the logical forms. The decoder also employs the multi-head attention mechanism. The decoder output is dependent on the encoder contextual embeddings $h$. The main task of the decoder is to generate each corresponding action, based on Table~\ref{tab:grammar}, alongside with the general semantic object from the knowledge graph (entity, type, predicate). In other words, the decoder will predict the main logical form without using or initialising any specific information from the knowledge graph. Here we define the decoder vocabulary as $V^{(dec)} = \{find, find\_reverse, \dots, entity, \allowbreak type, \allowbreak predicate, \allowbreak value\}$, where all the actions from Table~\ref{tab:grammar} are included. On top of the decoder stack, we employ a linear layer alongside a softmax to calculate each token's probability scores in the vocabulary. We define the decoder stack output as follows:
\begin{equation}
\small
\begin{split}
    &h^{(dec)} = decoder(h^{(enc)};\theta^{(dec)}),\\
    &p_{t}^{(dec)} = softmax(\boldsymbol{W}^{(dec)} h_{t}^{(dec)}),
\end{split}
\end{equation}
\noindent where $h_{t}^{(dec)}$ is the hidden state in time step $t$, $\theta^{(dec)}$ are the decoder trainable parameters, $\boldsymbol{W}^{(dec)} \in \mathbb{R}^{|V^{(dec)}|\times d}$ are the linear layer weights, and $p_{t}^{(dec)} \in \mathbb{R}^{|V^{(dec)}|}$ is the probability distribution over the decoder vocabulary in time step $t$. The $|V^{(dec)}|$ denotes the decoder's vocabulary size.
\subsection{Entity Recognition Module} \label{sec:er}
The entity recognition module is composed of two sub-modules, where each module is trained using a different objective. 
\subsubsection{Entity Detection and Linking}
\paragraph{Entity Detection}
It aims to detect and link the entities to the KG. The module is inspired by \cite{shen-etal-2019-multi} and performs type-aware entity detection by using BIO sequence tagging jointly with entity type tagging. Specifically, the entity detection vocabulary is defined as $V^{(ed)} = \{O, \{B, I\} \times \{TP_{i}\}_{i=1}^{N^{(tp)}}\}$, where $TP_{i}$ denotes the \textit{i-th} entity type label, $N^{(tp)}$ stands for the number of the distinct entity types in the knowledge graph and $|V^{(ed)}| = 2 \times N^{(tp)} + 1$.  For performing the sequence tagging task we use an LSTM \cite{hochreiter1997lstm} and the module is defined as:
\begin{equation}
\small
\begin{split}
    &h^{(l)} = LeakyReLU(LSTM(h^{(enc)};\theta^{(l)})), \\
    &p_{t}^{(ed)} = softmax(\boldsymbol{W}^{(l)} h_{t}^{(l)} ),
\end{split}
\end{equation}

\noindent where $h^{(enc)}$ is the encoder hidden state, $\theta^{(l)}$ are the LSTM layer trainable parameters, $h_{t}^{(l)}$ is the LSTM hidden state for time step $t$, $\boldsymbol{W}^{(l)} \in \mathbb{R}^{|V^{(ed)}| \times d}$ are the linear layer weights and $p_{t}^{(ed)}$ are the entity detection module prediction for time step $t$. $|V^{(ed)}|$ denotes the entity detection vocabulary size.

\paragraph{Entity Linking} 
Once the entity BIO labels and their types are recognised, the next steps for the entity linking are: 
1) the BIO labels are used to locate the entity spans from the input utterances. 2) An inverted index built for the knowledge graph entities is used to retrieve candidates for each predicted entity span. Finally, 3) the candidate lists are filtered using the predicted (entity) types. From the filtered candidates, the first entity is considered as correct.

\subsubsection{Filtering and Permutation}
After finding all the input utterances' entities, we perform two additional tasks in order to use entities in the generated logical form. First, we filter the relevant entities, and then we need to permute the entities in the order required for the logical form.
The module receives as an input the concatenation of the hidden states of the encoder $h^{(enc)}$ and the hidden states of the LSTM $h^{(l)}$ from the entity detection model. The module here learns to assign index tags to each input token. We define the module vocabulary as $V^{(ef)} = \{0, 1, \dots, m\}$ where $0$ is the index assigned to the context entities that are not considered. The remaining values are indices that permute our entities based on the logical form. Here, $m$ is the total number of indices based on the maximum number of entities from all logical forms. Overall, our filtering and permutation module is modelled using a feed-forward network with two linear layers separated with a Leaky ReLU activation function and appended with a softmax. Formally we define the module as:
\begin{equation}
\begin{split}
    &h^{(ef)} = LeakyReLU( \boldsymbol{W}^{(ef_1)} [h^{(enc)};h^{(l)}]), \\
    &p_{t}^{(ef)} = softmax( \boldsymbol{W}^{(ef_2)} h_{t}^{(ef)}),
\end{split}
\end{equation}
\noindent where $\boldsymbol{W}^{(ef_1)} \in \mathbb{R}^{d \times 2d}$ are the weights of the first linear layer and $h_{t}^{(ef)}$ is the hidden state of the module in time step $t$. $\boldsymbol{W}^{(ef_2)} \in \mathbb{R}^{|V^{(ef)}| \times d}$ are the weights of the second linear layer, $|V^{(ef)}|$ is the size of the vocabulary and $p_{t}^{(ef)}$ denotes the probability distribution over the tag indices for the time step $t$.

\subsection{Graph Attention-Based Module}
A knowledge graph (KG) can be denoted as a set of triples $\mathcal{K} \subseteq \mathcal{E} \times \mathcal{R} \times \mathcal{E}$ where  $\mathcal{E}$ and $\mathcal{R}$ are the set of entities and relations respectively. To build the (local) graph, we consider the relations and the types of entities that are linked with these relations in the knowledge graph $\mathcal{K}$. We define a graph $\mathcal{G} = \{\mathcal{T} \cup \mathcal{R}, \mathcal{L}\}$ where $\mathcal{T}$ is the set of types, $\mathcal{R}$ is the set of relations and $\mathcal{L}$ is a set of links $(tp_1, r)$ and $(r, tp_2)$ such that $\exists (e_1, r, e_2) \in \mathcal{K}$ where $e_1$ is of type $tp_1$ and $e_2$ is of type $tp_2$. 

To propagate information in the graph and to project prior KG information into the embedding space, we use the Graph Attention Networks (GATs) \cite{velickovic2018graph}. 

We initialise each node embedding $h^{(g)} = \{h_{1}^{(g)}, \dots, h_{n}^{(g)} \}$ using pretrained BERT embeddings, and $n = |\mathcal{T} \cup \mathcal{R}|$. A GAT layer uses a parameter weight matrix, and self-attention, to produce a transformation of input representations $\overline{h}^{(g)} = \{\overline{h}_{1}^{(g)}, \dots, \overline{h}_{n}^{(g)} \}$, where $\overline{h}_{i}^{(g)} \in \mathbb{R}
^{d}$ as shown below:
\begin{equation}
    \overline{h}^{(g)} = g(h^{(g)}; \theta^{(g)}), \footnote{For more details about GAT please refer to the appendix.}
\end{equation}
\noindent and $\theta^{(g)}$ are the trainable parameters. 
We model the task of predicting the correct type or predicate in the logical form as a classification task over the nodes in graph $\mathcal{G}$, given the current conversational context and decoder hidden state. For each time step $t$ in the decoder, we calculate the probability distribution $p_{t}^{(g)}$ over the graph nodes as: 
\begin{equation}
 p_{t}^{(g)} = softmax(\overline{h}^{(g)T} h_{t}^{(c)}),
\end{equation}
\noindent where $\overline{h}^{(g)} \in \mathbb{R}^{d \times n}$ and $h_{t}^{(c)}$ is a linear projection of the concatenation of the context representation and the decoder hidden state, given as follows, 
\begin{equation}
    h_{t}^{(c)} = LeakyReLU(\boldsymbol{W}^{(g)} [h_{ctx}^{(enc)};h_t^{(dec)}]),
\end{equation}
\noindent and $\boldsymbol{W}^{(g)} \in \mathbb{R}^{d \times 2d}$.
\subsection{Learning}
The framework consists of four trainable modules, grammar guided decoder, entity detection, filtering and permutation, and the GAT-based module for types and predicates. 
Every module consists of a loss function that contributes to the overall performance of the framework, as shown in Section~\ref{sec:ablation}. To account for multi-tasking, we perform a weighted average of all the single losses:
\begin{equation}
    L = \lambda_1 L^{dec} + \lambda_2 L^{ed} + \lambda_3 L^{ef} + \lambda_4 L^{g},
\end{equation}

\noindent where $\lambda_1, \lambda_2, \lambda_3, \lambda_4$ are the relative weights, which are learned during training by taking into account the difference in magnitude between losses by incorporating the log standard deviation \cite{armitage2020mlm,cipolla2018mltloss}. $L^{dec}, L^{ed}, L^{ef}, $ and $L^{g}$ are the respective negative log-likelihood losses of the grammar guided decoder, entity detection, filtering and permutation, and GAT-based modules. These losses are defined as follows: 

\begin{equation}
\begin{split}
    &L^{dec} = - \sum_{k=1}^{m} log p(y_{k}^{(dec)} | x), \\
    &L^{ed} = - \sum_{j=1}^{n} log p(y_{j}^{(ed)} | x), \\
    &L^{ef} = - \sum_{i=1}^{n} log p(y_{i}^{(ef)} | x), \\
    &L^{g} = - \sum_{k=1}^{m} I_{(y_{k}^{(dec)} \in\{type, pred\})} log p(y_{k}^{(g)} | x),
\end{split}
\end{equation}

\noindent where $n$ and $m$ are the length of the input utterance $x$ and the gold logical form, respectively. $y_{k}^{(dec)} \in V^{(dec)}$ are the gold labels for the decoder, $y_{j}^{(ed)} \in V^{(ed)}$ are the gold labels for entity detection,  $y_{j}^{(ef)} \in V^{(ef)}$ are the gold labels for filtering and permutation, and $y_{k}^{(g)} \in \{\mathcal{T} \cup \mathcal{R}\}$ are the gold labels for the GAT-based module. The model benefits from multiple supervision signals from each module, and this improves the performance in the given task. 

\begin{table*}[ht]
\centering
\resizebox{\textwidth}{!}{%
\begin{tabular}{c|c|cccc|c}
\toprule
\textbf{Methods} & & HRED-KVM & D2A & MaSP & LASAGNE (ours) & $\Delta$ \\
\textbf{\# Train Param} & & \multicolumn{1}{c}{-} & \multicolumn{1}{c}{-} & \multicolumn{1}{c}{15M} & \multicolumn{1}{c|}{14.7M} & \\ \hline
\textbf{Question Type} & \textbf{\#Examples} & \multicolumn{4}{c|}{F1 Score} \\ \hline
Overall & 206k & 9.39\% & 66.70\% & 79.26\% & \textbf{82.91\%} & +3.65\% \\
Clarification & 12k & 16.35\% & 35.53\% & \textbf{80.79\%} & 69.46\% & -11.33\% \\
Comparative Reasoning (All) & 15k & 2.96\% & 48.85\% & 68.90\% & \textbf{69.77\%} & +0.87\% \\
Logical Reasoning (All) & 22k & 8.33\% & 67.31\% & 69.04\% & \textbf{89.83\%} & +20.79\% \\
Quantitative Reasoning (All) & 9k & 0.96\% & 56.41\% & 73.75\% & \textbf{86.67\%} & +12.92\% \\
Simple Question (Coreferenced) & 55k & 7.26\% & 57.69\% & 76.47\% & \textbf{79.06\%} & +2.59\% \\
Simple Question (Direct) & 82k & 13.64\% & 78.42\% & 85.18\% & \textbf{87.95\%} & +2.77\% \\
Simple Question (Ellipsis) & 10k & 9.95\% & 81.14\% & \textbf{83.73\%} & 80.09\% & -3.64\% \\ \hline
\textbf{Question Type} & \textbf{\#Examples} & \multicolumn{5}{c}{Accuracy} \\ \hline
Overall & 66k & 14.95\% & 37.33\% & 45.56\% & \textbf{64.34\%} & +18.78\% \\
Verification (Boolean) & 27k & 21.04\% & 45.05\% & 60.63\% & \textbf{78.86\%} & +18.23\% \\
Quantitative Reasoning (Count) & 24k & 12.13\% & 40.94\% & 43.39\% & \textbf{55.18\%} & +11.79\% \\
Comparative Reasoning (Count) & 15k & 8.67\% & 17.78\% & 22.26\% & \textbf{53.34\%} & +31.08\% \\
\bottomrule
\end{tabular}%
}
\caption{LASAGNE's performance comparison on the CSQA dataset having 200K dialogues with 1.6M turns and over 12.8M entities. LASAGNE achieves ``overall'' (weighted average on all question types) new state of the art for both the F1 score and the question type results' accuracy metric.}
\label{tab:results}
\end{table*}

\section{Experiments} \label{sec:experiment} 
\subsection{Experimental Setup}
\paragraph{Datasets}
We use the Complex Sequential Question Answering (CSQA) dataset\footnote{\url{https://amritasaha1812.github.io/CSQA}} \cite{saha2018complex}. CSQA was built on the large-scale knowledge graph Wikidata. Wikidata consists of 21.2M triples with over 12.8M entities, 3,054 entity types, and 567 predicates. The CSQA dataset consists of around 200K dialogues where each partition -- train, valid, test contains 153K, 16K, 28K dialogues, respectively. The questions involve complex reasoning to determine the correct answers. 

\textbf{Model Configurations} 
We incorporate a semi-automated preprocessing step to annotate the CSQA dataset with gold logical forms. For each question type and subtype in the dataset, we create a general template with a pattern sequence that the actions should follow. Thereafter, we follow a set of rules to create the specific gold logical form that extracts the gold sequence of actions based on the type of question for each question. The actions used for this process are the ones in Table~\ref{tab:grammar}.
For all the modules in the LASAGNE framework, we employ an embedding dimension of 300. We utilise the transformer model with six heads for the multi-head attention model with two layers. For the optimisation, we use the Noam optimiser proposed by \cite{vaswani2017attention}, where authors use an Adam optimiser \cite{kingma2015adam} with several warmup steps for the learning rate. Please refer to the appendix submitted with the paper for more details.

\textbf{Models for Comparison}
We compare the LASAGNE framework with the last three baselines that have been evaluated on the employed dataset. The first baseline is \cite{saha2018complex} where authors introduce the HRED+KVmem model. The second baseline is D2A \cite{guo2018dialog}, which uses a semantic parsing approach based on a seq2seq model. Finally, the current state of the art is MaSP \cite{shen-etal-2019-multi}, which is also a semantic parsing approach. Please note, the number of parameters for LASAGNE were 14.7M compared to MaSP with 15M. Our base transformer model can be replaced with larger models like BERT with extremely large number of parameters for performance gain, however, that was not the focus of this work. 

\textbf{Evaluation Metrics}
We use the same metrics as employed by the authors of the CSQA dataset \cite{saha2018complex} as well as the previous baselines. The ``F1-score'' is used for questions that have an answer composed of a set of entities. The ``Accuracy'' metric is used for the question types whose answer is a number or a boolean value (YES/NO). We also provide an overall score for each evaluation metric and their corresponding question categories. 

\subsection{Results}
Table~\ref{tab:results} summarises the results comparing the LASAGNE framework against the previous baselines. LASAGNE outperforms the previous baselines weighted average on all question types (The row ``overall'' in the Table~\ref{tab:results}). Furthermore, LASAGNE is a new SotA in 8 out of 10 question types, and in some cases, the improvement is up to 31 percent.\\
\textbf{What worked in our case?} For question types that require more than two entities for reasoning, such as \textit{Logical Reasoning (All)} and \textit{Verification (Boolean)}, LASAGNE performs considerably better ($+20.79\%$ and $+18.23\%$ respectively). This is mainly due to the proposed entity recognition module. Furthermore, for question types that require two or more (entity) types and predicates, such as \textit{Quantitative Reasoning (All)}, \textit{Quantitative Reasoning (Count)} and \textit{Comparative Reasoning (Count)} LASAGNE also outperforms MaSP ($+12.92\%$, $+11.79\%$ and $+31.08\%$ respectively). Here, the improvement is due to the graph attention-based module, which is responsible for predicting the relevant (entity) types and predicates. Another interesting result is that LASAGNE also performs better in two out of three \textit{Simple Question involving one entity and one predicate} categories. The performance shows the robustness of LASAGNE.\\
\textbf{What did not work in our case?} 
LASAGNE noticeably under-performs on the \textit{Clarification} question type, where MaSP retains the state-of-the-art. The main reason is the spurious logical forms during the annotation process which has further impacted the Simple Questions (Ellipses) performance. 

\subsection{Ablation Study}\label{sec:ablation}
\textbf{Effect of GAT and Multi-task Learning}
Table~\ref{tab:ablation_study} summarises the effectiveness of the GAT-based module and the multi-task learning. We can observe the advantage of using them together in LASAGNE. 
To show the effectiveness of GAT-based module, we replace it with two simple classifiers, one for each predicate and type categories. We can observe that the performance drops significantly for the question types that require multiple entity types and predicates (e.g. \textit{Quantitative Reasoning (All)}, \textit{Quantitative Reasoning (Count)} and \textit{Comparative Reasoning (Count)}). 
When we exclude the multi-task learning and train all the modules independently, there is a negative impact on all question types. In LASAGNE, the filtering and permutation module, along with the GAT-based module, is heavily dependent on the supervision signals received from the previous modules. Therefore it is expected that without the multi-task learning, LASAGNE will underperform on all question types, since each module has to re-learn inherited information. 

\begin{table}[!t]
\centering
\resizebox{\textwidth}{!}{%
\begin{tabular}{c|ccc}
\toprule
\textbf{Methods} & Ours & w/o GATs & w/o Multi \\ \hline
\multicolumn{1}{c|}{\textbf{Question Type}} & \multicolumn{3}{c}{F1 Score} \\ \hline
\multicolumn{1}{c|}{Clarification} & 66.94\% & 57.33\% & 59.43\% \\
\multicolumn{1}{c|}{Comparative} & 69.77\% & 57.72\% & 66.41\% \\
\multicolumn{1}{c|}{Logical} & 89.83\% & 78.52\% & 86.75\% \\
\multicolumn{1}{c|}{Quantitative} & 86.67\% & 75.26\% & 82.18\% \\
\multicolumn{1}{c|}{Simple (Coref)} & 79.06\% & 76.46\% & 77.23\% \\
\multicolumn{1}{c|}{Simple (Direct)} & 87.95\% & 83.59\% & 85.39\% \\
\multicolumn{1}{c|}{Simple (Ellipsis)} & 80.09\% & 77.19\% & 78.47\% \\ \hline
\multicolumn{1}{c|}{\textbf{Question Type}} & \multicolumn{3}{c}{Accuracy} \\ \hline
\multicolumn{1}{c|}{Verification} & 78.86\% & 63.38\% & 75.24\% \\
\multicolumn{1}{c|}{Quantitative} & 55.18\% & 40.87\% & 46.27\% \\
\multicolumn{1}{c|}{Comparative} & 53.34\% & 41.73\% & 45.90\% \\
\bottomrule
\end{tabular}
}
\caption{The effectiveness of the GAT and the multi-task learning. The first column contains the results of the LASAGNE framework, where all the modules are trained simultaneously. The second and third columns selectively remove the GAT and the multi-task learning from LASAGNE.}
\label{tab:ablation_study}
\end{table}

\subsection{Task Analysis}

\begin{table}[ht]
\small
\centering
\begin{tabular}{c|c}
\toprule
Tasks & Accuracy \\ \hline
Entity Detection & 86.75\% \\
Filtering \& Permutation& 97.49\% \\
Grammar-Guided Decoder for Logical Forms & 98.61\% \\
GAT-Based Module for Type/Predicate & 92.28\% \\
\bottomrule
\end{tabular}
\caption{Tasks accuracy of the LASAGNE framework.}
\label{tab:task_analysis}
\end{table}

Table~\ref{tab:task_analysis} illustrates the task accuracy of LASAGNE. The Entity Detection task has the lowest accuracy ($86.75\%$). The main reason here is the errors in the entity type prediction. On the other hand, for all other tasks, we have accuracy above $90\%$.

\textbf{Effect of Filtering and Permutation }
For justifying the effectiveness and superior performance of LASAGNE's filtering and permutation module, we compare the overall performance of the entity recognition module to the corresponding module from MaSP. Please note, entity detection modules in both frameworks adopt a similar approach as defined in section \ref{sec:er}. In Table~\ref{tab:ablation_filtering_permutation} we can see that the MaSP entity recognition module provides an overall accuracy of $79.8\%$ on test data, while our module outperforms it with an accuracy of $92.1\%$. The main reason for the under-performance of MaSP is that it uses only token embeddings without any entity information. In contrast, our approach avoids re-learning entity information in the question context and relies on the entity detection module's information. 

\begin{table}[ht]
\small
\centering
\begin{tabular}{c|c}
\toprule
Model & Entity Recognition Accuracy \\ \hline
MaSP & 79.8\% \\
LASAGNE & 92.1\% \\
\bottomrule
\end{tabular}
\caption{Comparing MaSP~\cite{shen-etal-2019-multi} and LASAGNE for entity recognition performance.}
\label{tab:ablation_filtering_permutation}
\end{table}


\subsection{Error Analysis}
For the error analysis, we randomly sampled 100 incorrect predictions. We detail the reasons for two types of errors observed in the analysis:

\textbf{Entity Ambiguity}
Even though our entity detection module assigns (entity) types to each predicted span, entity ambiguity remains the biggest challenge for our framework. 
For instance, for the question, ``Who is associated with Jeff Smith ?'' LASAGNE entity detection module correctly identifies ``Jeff Smith'' as an entity surface form and correctly assigns the (entity) type ``common name''. However, the Wikidata knowledge graph contains more than ten entities with exactly the same label and type. Our entity linking module has difficulties in such cases. Wikidata entity linking is a newly emerging research domain that has its specific challenges such as entities sharing the same labels, user-created non-standard entity labels and multi-word entity labels (up to 62 words) \cite{Mulang2019ContextawareEL}. Additional entity contexts, such as entity descriptions and other KG contexts, could help resolve the Wikidata entity ambiguity \cite{mulang2020evaluating}.

\textbf{Spurious Logical Form}
For specific question categories, we could not identify gold actions for all utterances. Therefore spurious logical form is a standard error that affects LASAGNE. Specifically, we have spurious logical forms for categories such as ``Comparative, Quantitative, and Clarification'' but still can achieve SotA in the comparative and quantitative categories.


\section{Conclusions} \label{sec:conclusion}
In this article, we focus on complex question answering over a large-scale knowledge graph containing conversational context. We provide a transformer-based framework to handle the task in a multi-task semantic parsing manner. At the same time, we propose a named entity recognition module for entity detection, filtering, and permutation. 
Furthermore, we also introduce a graph attention-based module, which exploits correlations between (entity) types and predicates for identifying the gold ones for each particular context. 
We empirically show that our model achieves the best results for numerous question types and also overall. 
Our ablation study demonstrates the effectiveness of the multi-task learning and of our graph-based module. We also present an error analysis on a random sample of ``wrong examples'' to discuss our model's weaknesses. 
For future work, we believe that reinforcement learning is a viable alternative to explore complex conversational question answering without gold annotations. 

\section*{Acknowledgments}
The project leading to this publication has received funding from the European Union’s Horizon 2020 research and innovation program under the Marie Skłodowska-Curie grant agreement No.~812997 (Cleopatra).

\bibliographystyle{acl_natbib}
\bibliography{main.bbl}

\appendix

\section{Grammar}
We propose a new grammar to annotate the dataset with a gold logical form to perform the semantic parsing task. We consider the work by \cite{guo2018dialog} as a starting point for generating them. While we differ in many actions regarding their semantic and therefore their implementation. Our goal was to define more precise and richer actions, which gives us a more flexible grammar in terms of being used to annotate a wider range of question's complexities. For instance, for a couple of actions, we also define their reverse occurrence (e.g. \emph{find}, \emph{find\_reverse)}).
We do this in order to match the knowledge graph triple direction (\textit{subject-predicate-object}). In some questions, we might have the subject or the object entity. Having both normal and reverse actions helps us to identify directly the correct answer based on the action the model predicted. Furthermore, we also define actions that do not exist in \cite{guo2018dialog}. Some of them are \textit{find\_tuple\_counts}, \textit{atmost}, \textit{atleast}. Table~\ref{tab:grammar} illustrates the complete grammar with all the defined actions. Following \cite{lu2008generative}, we define each action with a function that can be executed on the knowledge graph. Finally, in order to execute a sequence of actions, we have to parse it into a tree structure. There our executor starts from the tree leaves and it recursively executes the leftmost non-terminal node until the whole tree is complete. 

\section{Case Study}
Table~\ref{tab:lf_examples} shows examples from different question types in the CSQA dataset and the logical forms generated from our model. As we can see, our actions can cover reasoning for every question type by following certain patterns depending on them. The sequences can cover all the different complexities of the questions. For example, the logical forms pattern of \emph{Simple Questions} to \emph{Quantitative} or \emph{Comparative} is slightly different due to increased complexity of the latter questions. The reasoning over \emph{Quantitative} or \emph{Comparative} question involves more actions in order to reach the correct answer. 

For the question type \emph{Simple Question (Direct)}, we can see the question ``Which administrative territory is the birthplace of Antonio Reguero ?''. The correct logical form for this example is ``filter\_type(find(Antonio Reguero, place of birth), administrative territorial entity)''. Here we can distinguish two different actions; the first one is the \textit{filter\_type} and the other one is the \textit{find} action. The \textit{find} action receives as input an entity subject and a predicate and provides the set of object entities from the Knowledge Graph. Whereas, the \textit{filter\_type} action receives as input a set of entities along with an entity type and results to a set of entities that belong to that particular entity type.

\section{Hyperparamters and module configurations}

\begin{table}[!t]
\small
\centering
\begin{tabular}{cc}
\toprule
\textbf{Hyperparameters} & \textbf{Value} \\ \hline
epochs & 20 \\
batch size & 64 \\
dropout ratio & 0.1 \\
learning rate & 0.001 \\
warmup steps & 4000 \\
optimizer & Adam \\
$\beta_1$ & 0.9 \\
$\beta_2$ & 0.999 \\
$\varepsilon$ & 1e-09 \\
model dimension & 300 \\
model pretrained embeddings & GloVe \\
non-linear activation & LeakyRelu \\
GAT input dimension & 3072 \\
GAT node dimension & 300 \\
GAT pretrained embeddings & BERT \\
\bottomrule
\end{tabular}
\caption{Hyper-parameters for LASAGNE framework.}
\label{tab:tab_hyp1}
\end{table}

\begin{table*}[!ht]
\centering
\resizebox{\textwidth}{!}{%
\begin{tabular}{c|c|cc|cc|cc|cc}
\toprule
\textbf{Methods} &  & \multicolumn{2}{c|}{HRED-KVM} & \multicolumn{2}{c|}{D2A} & \multicolumn{2}{c|}{MaSP} & \multicolumn{2}{c}{LASAGNE (ours)} \\ \hline
\textbf{Question Type} & \textbf{\#Examples} & Precision & Recall & Precision & Recall & Precision & Recall & Precision & Recall \\ \hline
Overall & 206k & 6.30\% & 18.40\% & 66.57\% & 66.83\% & 80.48\% & 78.07\% & 87.08\% & 80.31\% \\
Clarification & 12k & 12.13\% & 25.09\% & 33.97\% & 37.24\% & 77.66\% & 84.18\% & 81.80\% & 60.35\% \\
Comparative Reasoning (All) & 15k & 4.97\% & 2.11\% & 54.68\% & 44.14\% & 81.20\% & 59.83\% & 83.88\% & 59.73\% \\
Logical Reasoning (All) & 22k & 5.75\% & 15.11\% & 68.86\% & 65.82\% & 78.00\% & 61.92\% & 98.67\% & 82.43\% \\
Quantitative Reasoning (All) & 9k & 1.01\% & 0.91\% & 60.63\% & 52.74\% & 79.02\% & 69.14\% & 79.66\% & 95.02\% \\
Simple Question (Coreferenced) & 55k & 5.09\% & 12.67\% & 56.94\% & 58.47\% & 76.01\% & 76.94\% & 72.71\% & 86.62\% \\
Simple Question (Direct) & 82k & 8.58\% & 33.30\% & 77.37\% & 79.50\% & 84.29\% & 86.09\% & 94.94\% & 81.92\% \\
Simple Question (Ellipsis) & 10k & 6.98\% & 17.30\% & 77.90\% & 84.67\% & 82.03\% & 85.50\% & 93.74\% & 69.91\% \\ \hline
\textbf{Question Type} & \textbf{\#Examples} & \multicolumn{8}{c}{Accuracy} \\ \hline
Overall & 66k & \multicolumn{2}{c|}{14.95\%} & \multicolumn{2}{c|}{37.33\%} & \multicolumn{2}{c|}{45.56\%} & \multicolumn{2}{c}{64.34\%} \\
Verification (Boolean) & 27k & \multicolumn{2}{c|}{21.04\%} & \multicolumn{2}{c|}{45.05\%} & \multicolumn{2}{c|}{60.63\%} & \multicolumn{2}{c}{78.86\%} \\
Quantitative Reasoning (Count) & 24k & \multicolumn{2}{c|}{12.13\%} & \multicolumn{2}{c|}{40.94\%} & \multicolumn{2}{c|}{43.39\%} & \multicolumn{2}{c}{55.18\%} \\
Comparative Reasoning (Count) & 15k & \multicolumn{2}{c|}{8.67\%} & \multicolumn{2}{c|}{17.78\%} & \multicolumn{2}{c|}{22.26\%} & \multicolumn{2}{c}{53.34\%} \\
\bottomrule
\end{tabular}%
}
\caption{Precision and recall comparison with baselines.}
\label{tab:results_pr}
\end{table*}

Table~\ref{tab:tab_hyp1} summarizes the hyperparameters used across the LASAGNE framework.
For the transformer module, we use the configurations from \cite{vaswani2017attention}. Our model dimension is $d_{model} = 300$, with a total number of $H=6$ heads and $L=2$ layers. The inner feed-forward linear layers have dimension $d_{ff}  = 600$. Following the base transformer parameters, we apply residual dropout \cite{srivastava2014dropout} to the summation of the embeddings and the positional encodings in both encoder and decoder stacks with a rate of $0.1$. 
The entity detection module has a dimension of $300$. Our base LSTM here is followed with a LeakyReLU, dropout, and a linear layer. The output of the linear layer is the module prediction while the LSTM hidden state is propagated to the filtering and permutation layer.
The filtering and permutation module receives an input of dimension $600$ where here a linear layer is responsible to reduce it to $300$ which is the framework dimension. Like in the previous module, a LeakyReLU, dropout, and a linear layer are used for the final predictions. 
Finally, for the GAT-based module, we use pre-trained BERT embeddings for type and predicate labels. Hence the input dimension on this module is $3072$. The GAT layer will produce representations with an embedding size of $300$. Next, multiple linear, dropout, and LeakyReLU layers are used to produce the final predictions.

\section{Graph Attention Networks}

\begin{figure}[!t]
\centering
\captionsetup{type=figure}
\includegraphics[width=0.85\textwidth]{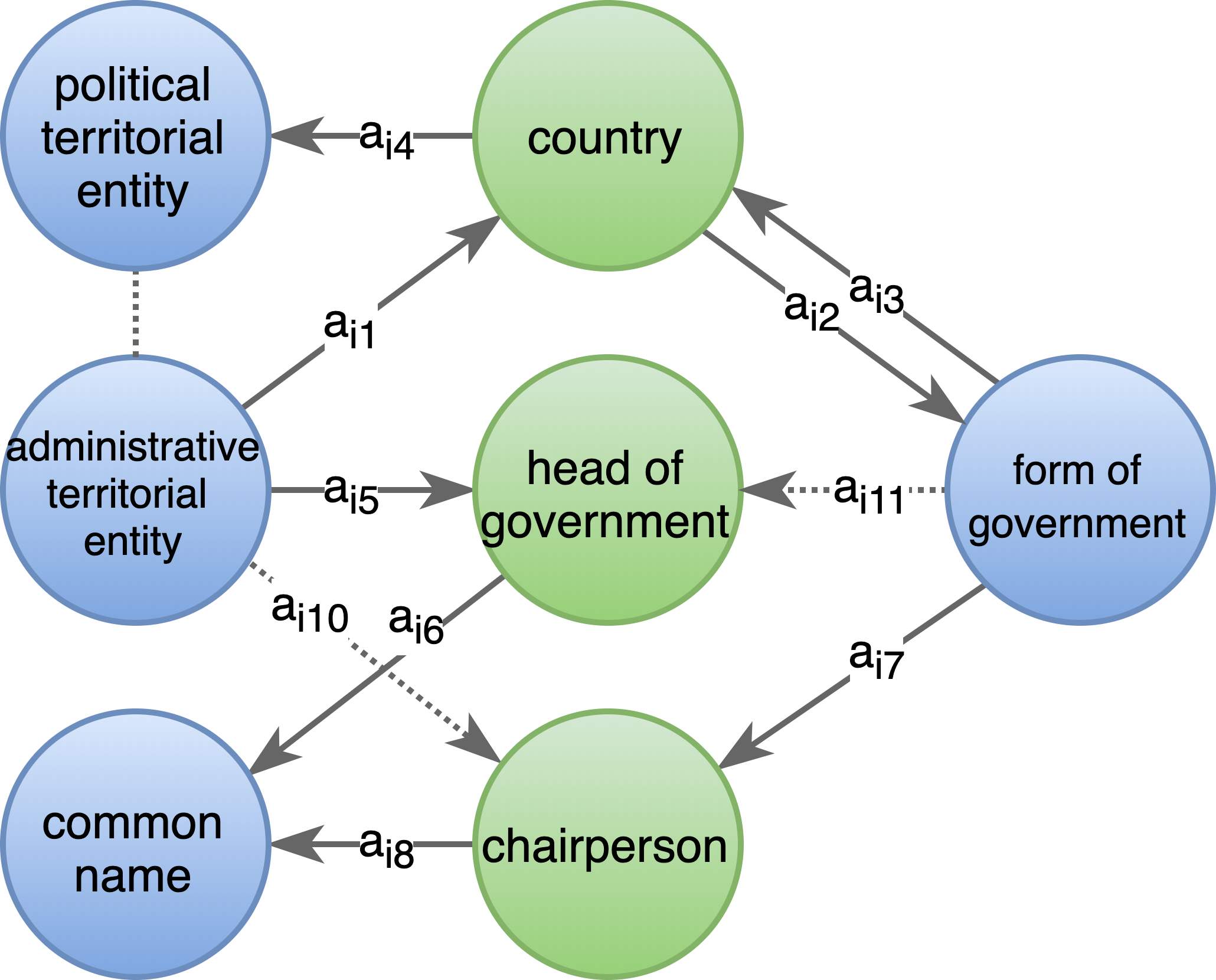}
\caption{The aggregation process of graph attention layer between the (entity) types and predicates from Wikidata knowledge graph. The dashed lines represent an auxiliary edge, while $a_{ij}$ represents relative attention values of the edge. We also incorporate the predicates (relations) as nodes of the graph instead of edges.}
\label{fig:gat_layer}
\end{figure}
Figure \ref{fig:gat_layer} shows the aggregation process of graph attention layer between the (entity) types and predicates from Wikidata. The KB types and predicates are the nodes of the graph, and there exist an edge only between types and predicates with the condition that there exist a triple which involved the predicate and an entity of that type. We use GATs \cite{velickovic2018graph} to capture different level of information for a node, based on the neighborhood in the graph.  We denote with $h^{(g)} = \{h_{1}^{(g)}, \dots, h_{n}^{(g)} \}$ the initial representations of the nodes, which will also be the input features for the GAT layer. To denote the influence of node $j$ to the node $i$, an attention score $e_{ij}$ is computed as $e_{ij} = a( \mathbf{W} h_{i}^{(g)}, \mathbf{W} h_{j}^{(g)}),$ where $\mathbf{W}$ is a parameterized linear transformation, and $a$ is an attention function. In our case, we follow the GAT paper, and compute $ e_{ij}$ score as follows,

\begin{equation}
    e_{ij} = LeakyReLU( \mathbf{a}^{T}[\mathbf{W} h_{i}^{(g)} || \mathbf{W} h_{j}^{(g)}]),
\end{equation}

\noindent where $\mathbf{a} \in \mathbb{R}^{2d}$ is a single-layer feedforward network, and  $||$ denotes concatenation. This attention scores are normalized using a softmax function and producing the $\alpha_{ij}$ scores for all the edges in a neighborhood. These normalized attention scores are used to compute the output features $\overline{h}_{i}^{(g)}$ of a node in a graph, by applying a linear combination of all the nodes in the neighborhood as below, 

\begin{equation}
    \overline{h}_{i}^{(g)} = \sigma ( \sum_{j \in \mathcal{N}_{i}} \alpha_{ij} \mathbf{W}  h_{j}^{(g)})
\end{equation}

\noindent where $\sigma$ is a non-linear function. 
Following \cite{velickovic2018graph} and \cite{vaswani2017attention} we also apply a multi-head attention mechanism and compute the final output features as, 

\begin{equation}
    \overline{h}_{i}^{(g)} = \sigma ( \frac{1}{K} \sum_{k=1}^{K} \sum_{j \in \mathcal{N}_{i}} \alpha_{ij}^{k} \mathbf{W}^{k}  h_{j}^{(g)})
\end{equation}

\noindent where $K$ is equal to the number of heads, and $\alpha_{ij}^{k}$, $\mathbf{W}^{k}$ are the corresponding attention scores and linear transformation by the $k$-th attention mechanism. During our experiments, we found out the $K=2$ was sufficient for our model. 

\section{Experiments}
Table~\ref{tab:results_pr} summarizes precision and recall results comparing LASAGNE framework against the previous baselines.
Furthermore, a detailed task analysis for each task on each question type is illustrated on Table~\ref{tab:tasks_analysis}.

\begin{table*}
\small
\centering
\begin{tabular}{c|cccc}
\toprule
\textbf{Tasks} & Entity Detection & Filt. \& Permut. & Logical Form & Type/Predicate \\ \hline
\textbf{Question Type} & \multicolumn{4}{c}{Accuracy} \\ \hline
Clarification & 92.19\% & 99.97\% & 98.36\% & 86.70\% \\
Comparative Reasoning (All) & 92.03\% & 99.88\% & 99.00\% & 97.18\% \\
Logical Reasoning (All) & 72.20\% & 99.44\% & 98.18\% & 95.95\% \\
Quantitative Reasoning (All) & 87.38\% & 100.0\% & 99.56\% & 95.87\% \\
Simple Question (Coreferenced) & 93.50\% & 96.92\% & 98.50\% & 90.12\% \\
Simple Question (Direct) & 90.58\% & 99.34\% & 98.58\% & 89.71\% \\
Simple Question (Ellipsis) & 77.90\% & 99.98\% & 98.81\% & 90.02\% \\
Verification (Boolean) & 79.50\% & 84.66\% & 99.79\% & 98.10\% \\
Quantitative Reasoning (Count) & 77.77\% & 99.80\% & 97.34\% & 92.09\% \\
Comparative Reasoning (Count) & 92.04\% & 99.98\% & 98.66\% & 96.92\% \\
\bottomrule
\end{tabular}
\caption{Task accuracy from LASAGNE. We can obtain that entity detection is the task with lowest accuracy while filtering and permutation together with logical form generation are the tasks with highest accuracy.}
\label{tab:tasks_analysis}
\end{table*}

\begin{table*}[!t]
\centering
\def\arraystretch{1.2}
\resizebox{\textwidth}{!}{%
\begin{tabular}{lll}
\toprule
\textbf{Question Type} & \textbf{Question} & \textbf{Logical Forms} \\ \hline
\begin{tabular}[c]{@{}l@{}}Simple \\ (Direct)\end{tabular} & \begin{tabular}[c]{@{}l@{}}Q1: Which administrative territory is the \\ birthplace of Antonio Reguero ?\end{tabular} & \begin{tabular}[c]{@{}l@{}}filter\_type(\\ \MyIndent find(Antonio Reguero, place of birth), \\ administrative territorial entity)\end{tabular} \\ \hline
\begin{tabular}[c]{@{}l@{}}Simple \\ (Ellipsis)\end{tabular} & \begin{tabular}[c]{@{}l@{}}Q1: Which administrative territories are \\ twin towns of Madrid ?\\ A1: Prague, Moscow, Budapest\\ Q2: And what about Urban \\ Community of Brest?\end{tabular} & \begin{tabular}[c]{@{}l@{}}filter\_type( \\ \MyIndent find(Urban Community of Brest, twinned administrative body), \\ administrative territorial entity)\end{tabular} \\ \hline
\begin{tabular}[c]{@{}l@{}}Simple \\ (Coref)\end{tabular} & \begin{tabular}[c]{@{}l@{}}Q1: What was the sport that Marie Pyko \\ was a part of ?\\ A1: Association football\\ Q2: Which political territory does that \\ person belong to ?\end{tabular} & \begin{tabular}[c]{@{}l@{}}filter\_type(\\ \MyIndent find(Marie Pyko, country of citizenship),\\ political territorial entity)\end{tabular} \\ \hline
\begin{tabular}[c]{@{}l@{}}Quantitative \\ Reasoning \\ (Count)\end{tabular} & \begin{tabular}[c]{@{}l@{}}Q1: How many beauty contests and business \\ enterprises are located at that city ?\\ A1: Did you mean Caracas?\\ Q2: Yes\end{tabular} & \begin{tabular}[c]{@{}l@{}}count(union(\\ \MyIndent filter\_type(find\_reverse(Caracas, located in), beauty contest), \\ \MyIndent filter\_type(find\_reverse(Caracas, located in), business enterprises)
\\))\end{tabular} \\ \hline
\begin{tabular}[c]{@{}l@{}}Quantitative \\ Reasoning \\ (All)\end{tabular} & \begin{tabular}[c]{@{}l@{}}Q1; Which political territories are known to \\ have diplomatic connections \\ with max number of political territories ?\end{tabular} & \begin{tabular}[c]{@{}l@{}}argmax(\\ \MyIndent find\_tuple\_counts(diplomatic relation, political territorial entity, \\ \MyIndent political territorial entity))\end{tabular} \\ \hline
\begin{tabular}[c]{@{}l@{}}Comparative \\ Reasoning \\ (Count)\end{tabular} & \begin{tabular}[c]{@{}l@{}}Q1: How many alphabets are used as the \\ scripts for more number of languages \\ than Jawi alphabet ?\end{tabular} & \begin{tabular}[c]{@{}l@{}}count(greater(count(\\ \MyIndent filter\_type(find(Jawi alphabet, writing system), language)), \\ \MyIndent find\_tuple\_counts(writing system, alphabet, language)))\end{tabular} \\ \hline
\begin{tabular}[c]{@{}l@{}}Comparative \\ Reasoning \\ (All)\end{tabular} & \begin{tabular}[c]{@{}l@{}}Q1: Which occupations were more number \\ of publications and works mainly \\ about than composer ?\end{tabular} &
\begin{tabular}[c]{@{}l@{}}greater(union(\\ \MyIndent find\_reverse\_tuple\_counts(main subject, occupation, publication), \\ \MyIndent find\_reverse\_tuple\_counts(main subject, occupation, work)), \\ \MyIndent count(filter\_multi\_types(find\_reverse(composer, main subject), {publication, work})))\end{tabular} \\ \hline
Verification & \begin{tabular}[c]{@{}l@{}}Q1: Was Geir Rasmussen born at that \\ administrative territory ?\end{tabular} & is\_in(find(Geir Rasmussen, place of birth), Chicago) \\
\bottomrule
\end{tabular}
}
\caption{Examples from the CSQA dataset \cite{saha2018complex} annotated with gold logical form.}
\label{tab:lf_examples}
\end{table*}

\end{document}